\documentclass[letterpaper, 10 pt, conference]{ieeeconf}  

\IEEEoverridecommandlockouts                              

\usepackage[ruled, linesnumbered]{algorithm2e}
\usepackage{color}

\usepackage{graphicx} 
\usepackage{amsmath} 
\usepackage{multirow}
\usepackage{subfigure}

\usepackage{comment}

\title{\LARGE \bf
osmAG: Hierarchical Semantic Topometric Area Graph Maps in the OSM Format for Mobile Robotics  
}

\usepackage{hyperref}

\author{Delin Feng, Chengqian Li, Yongqi Zhang, Chen Yu, and S\"oren Schwertfeger
\thanks{Delin Feng, Chengqian Li, Yongqi Zhang, Chen Yu, and S\"{o}ren Schwertfeger are with the
	Key Laboratory of Intelligent Perception and Human-Machine Collaboration (ShanghaiTech University), Ministry of Education, Shanghai, China
        {\tt\small \{fengdl, lichq, zhangyq, yuchen, soerensch\}@shanghaitech.edu.cn}}%
}

\begin{document}

\maketitle
\thispagestyle{empty}
\pagestyle{empty}

\begin{abstract}
Maps are essential to mobile robotics tasks like localization and planning. We propose the open street map (osm) XML based Area Graph file format to store hierarchical, topometric semantic multi-floor maps of indoor and outdoor environments, since currently no such format is popular within the robotics community. Building on-top of osm we leverage the available open source editing tools and libraries of osm, while adding the needed mobile robotics aspect with building-level obstacle representation yet very compact, topometric data that facilitates planning algorithms. Through the use of common osm keys as well as custom ones we leverage the power of semantic annotation to enable various applications. For example, we support planning based on robot capabilities, to take the locomotion mode and attributes in conjunction with the environment information into account. The provided C++ library is integrated into ROS. We evaluate the performance of osmAG using real data in a global path planning application on a very big osmAG map, demonstrating its convenience and effectiveness for mobile robots.

\end{abstract}

\section{INTRODUCTION}
Autonomous mobile robots have found extensive applications in various fields such as agriculture, industry, and commerce. These robots rely on known maps for autonomous localization, mission-planning and navigation to accomplish tasks. A well-defined map representation serves as a prerequisite for mobile robots to execute their tasks effectively. Given the substantial differences in scale and structural characteristics between indoor and outdoor environments, map representations for mobile robots vary significantly. For outdoor scenarios, large-scale vector maps like OpenStreetMap are common. Point cloud maps are often employed in smaller outdoor scenarios \cite{kostavelis2015semantic}. In indoor settings, 2D or 3D grid maps and point cloud maps are the most widely used map representations. Consequently, most robots are designed for either indoor or outdoor environments exclusively.

With the further advancement of sensor technology, maps are becoming more accurate and detailed. Simultaneously, improvements in simultaneous localization and mapping (SLAM) algorithms have led to increased precision and scalability in map construction. However, memory and computation constrains limit the use of point cloud and 3D grid map based approaches for large areas for localization. They are also typically too detailed for most semantic information that is to be encoded, like room numbers or terrain types. Finally, 3D mesh and point cloud based maps are quite difficult to employ for path planning and navigation. Therefore, for most of these tasks, 2D grid maps are still widely used in robotics. They are suitable up to a certain size in terms of storage and computation, but become computationally more demanding as map sizes grow. They do not readily provide semantic information and, most crucially, are inherently 2D, so fail to support multi-level scenarios like buildings with more than one floor.

\begin{figure}[t]
	\centering    \includegraphics[width=0.45\textwidth]{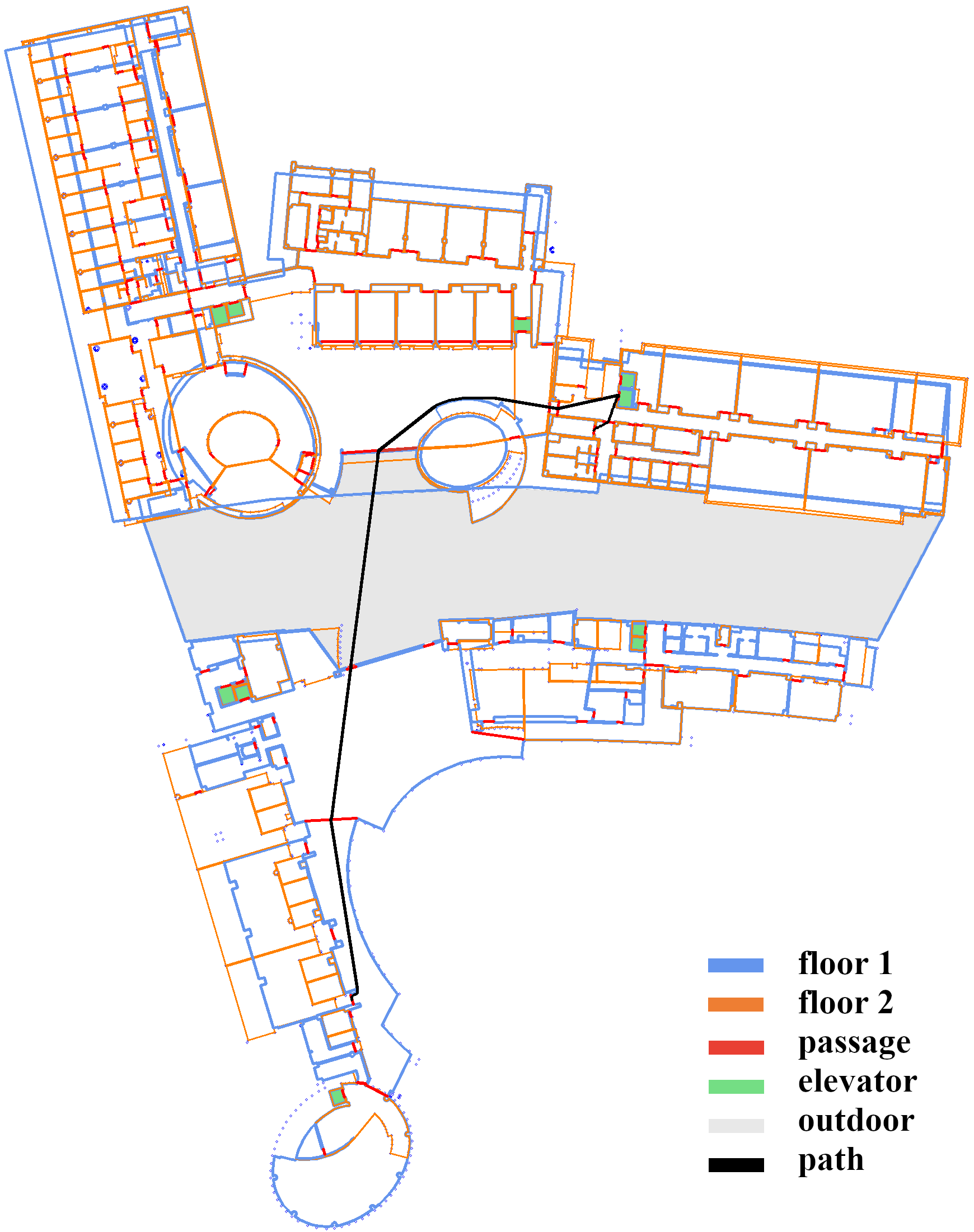}
	\caption{Two buildings with two stories each rendered from osmAG and with a planned path utilizing an elevator from one building to the other. }
	\label{figurelabe0}
\end{figure}

Graph structures encoding the topology of an environment are a solution to these problems \cite{Thrun2003RoboticMA}. These graphs represent places (vertices) and how they are connected (edges). Topometric maps additionally encode metric information about the position of these elements. These graphs can also be easily organized in a hierarchical manner, grouping several places or connections into a parent structure. Adding semantic information to the graph elements further enhances the utility of the map. In our paper \cite{he2021hierarchical} we go further into details about hierarchical topometric representation of 3D robotic maps. 

In our previous work \cite{Hou2019AreaGraph} and \cite{hou2022matching} we developed the so-called Area Graph, a topometric representation generated from 2D grid maps. That work is based on a so-called TopologyGraph~\cite{schwertfeger2016map}, which is a pruned and simplified version of the Voronoi Diagram. The Area Graph has as graph nodes areas and as edges passages between the areas: passages mark where it is possible for a robot to traverse from one area to another area that is touching the first area. \\

In this paper we present an on-file format to store said Area Graph. Additionally, we define how a hierarchy of areas and passages can be established and how semantic information can be added to it. That on-file format is based on and fully compatible with the open street map XML format - we just define certain use patterns and special osm tags that encode the hierarchical, semantic Area Graph structure in the osm XML tag ``way", that is used for both areas and passages. We also provide the osmAG C++ library to load osmAG files into memory and utilize them for visualization and planning in the Robot Operating System (ROS). Fig. \ref{figurelabe0} shows a multi-story osmAG map and a planned path.\\

Adopting the open street map XML standard has advantages in its widespread use and support from mature software, facilitating map editing, visualization, updates, and corrections. Apart from autonomous cars and a few other notable exceptions \cite{hentschel2010autonomous}, open street map has not been widely adopted in the robotics community, even though some indoor approaches exist \cite{naik2019semantic, wang2018data}. \\

The AreaGraph encodes the areas of rooms based on the architectural walls. We can create the AreaGraph from complete 3D point clouds \cite{he2021hierarchical} or from 2D grid maps  \cite{Hou2019AreaGraph, hou2022matching}. We obtain furniture-free 2D grid maps by either employing according to Simultaneous Localization and Mapping technologies \cite{He2019FurnitureFree, feng2023Floorplannet} or by rendering CAD building data into 2D grid maps. The memory consumption of our abstract osmAG is orders of magnitude smaller than 2D or 3D grid maps and point clouds, and planning on them is easy and fast, as our experiment will show. In  \cite{xie2023robust} we already utilize osmAG for localization and tracking of a mobile robot with a 3D LiDAR, see Fig. \ref{fig:localization}.\\

The key contributions of this work are: 

\begin{itemize}
 \item Definition of an on-disk map representation osmAG to store hierachical, semantic, topometric map data in the open street map format with additional robotics-centric Area Graph data.
 \item Providing a C++ library osmAGlib to load osmAG into memory and perform tasks like visualization in ROS and path planning. 
  \item Providing a big osmAG map and performing various experiments with it \footnote{\url{https://github.com/STAR-Center/osmAG}}.
\end{itemize}


\section{Related work}

\subsection{Hierarchical Topological Maps}

In recent years, there has been great interest in the integration of hierarchical, semantic, and topological mapping.
Specifically, learning-based semantic segmentation has made significant progress, enabling high-precision semantic segmentation of 3D point clouds and images. Current research on semantic maps mainly focuses on object-level
 \cite{semanticslam, QuadricSLAM}
  or dense maps, which include volumetric models 
\cite{SemanticFusion},
 point clouds 
 \cite{SemanticKITTI, denseslam},
  and 3D grids (voxels) \cite{Kimera}. These methods do not involve the estimation of higher-level semantics, such as rooms, and often return dense models that are not suitable for direct navigation. Segmentation alone also does not provide topological connections of neighboring areas.

\begin{figure}[t]
	\centering
	\includegraphics[width=0.49\textwidth]{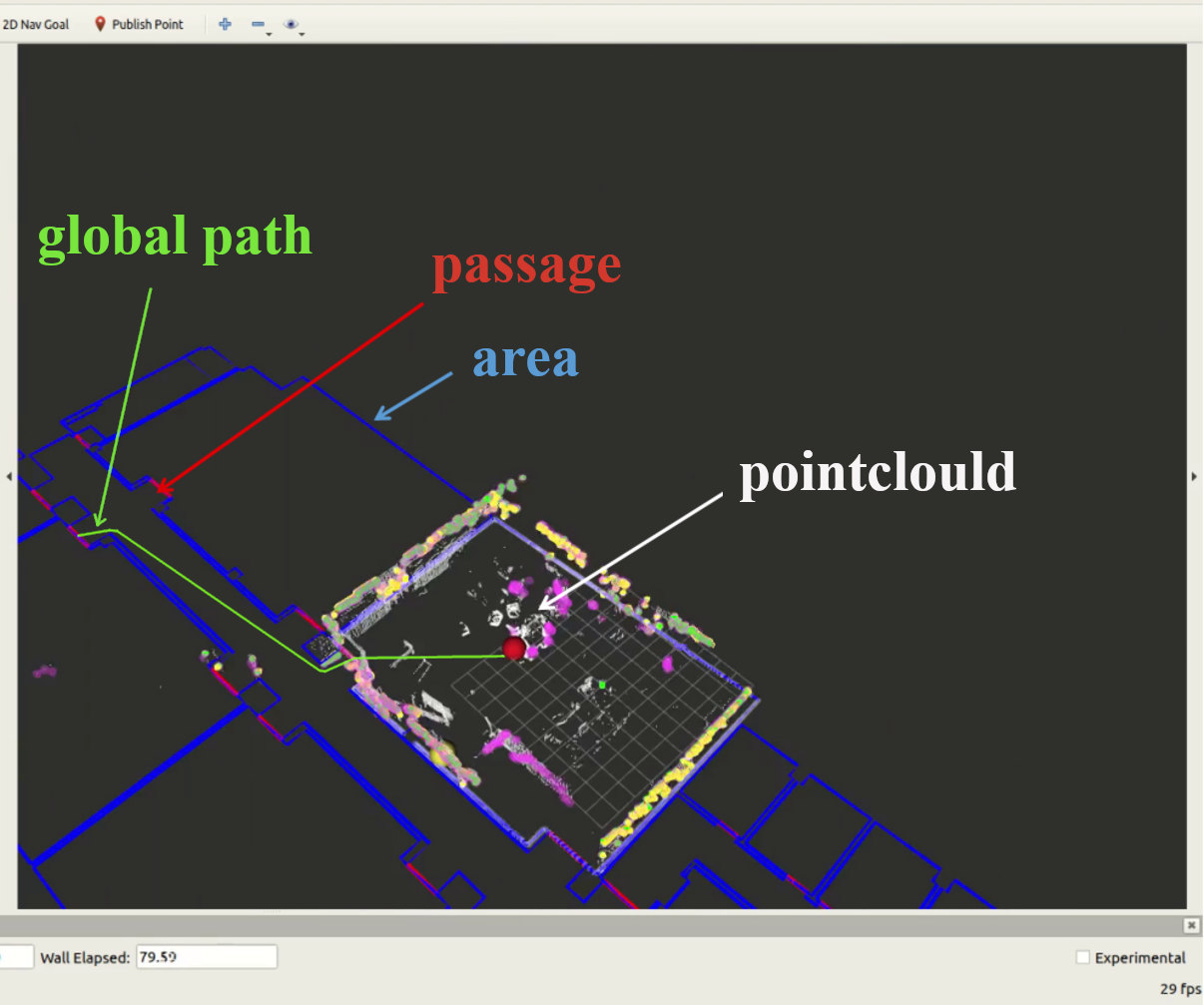}
	\caption{Application of osmAG for robot localization with a 3D LiDAR pointcloud in the Robot Operating System ROS \cite{xie2023robust} }
	\label{fig:localization}
\end{figure}

The second research line focuses on constructing hierarchical topological map models. Hierarchical maps have been widely used since the inception of robotics 
\cite{Thrun2003RoboticMA, KUIPERS1978129}.
Early works focus on 2D maps, exploring the use of multi-layer maps to address the apparent discrepancy between metric and topological representations \cite{RUIZSARMIENTO2017257, Multi-hierarchical}. 

More recently, 3D Scene Graphs have been introduced as expressive layered models for 3D environments. A 3D Scene Graph is a hierarchical graph where nodes represent spatial concepts at multiple abstraction levels
and edges represent relationships between concepts. Armeni et al. \cite{3DSceneGraph} pioneer the use of 3D Scene Graphs in computer vision, modeling the environment as a graph that includes low-level geometry (i.e., metric-semantic grids) , objects, rooms, and camera poses, and presented the first algorithm to parse metric-semantic 3D grids into 3D Scene Graphs. 

Rosinol et al. \cite{rosinol20203d} propose a novel 3D Scene Graph model that directly builds from sensor data, including subgraphs for places (useful for robot navigation), modeling objects, rooms, and buildings, and capturing moving entities in the environment. 

The work most similar to ours is a semantic SLAM method Hydra \cite{hughes2022hydra}. Hydra constructs a real-time multi-level 3D Scene Graph using onboard sensors. It transforms a local Euclidean Signed Distance Function (ESDF) into a metric-semantic 3D grid and a Voronoi diagram. It extracts a topological map of places and uses a method inspired by community detection for room segmentation. Our proposed osmAG has a similar hierarchical structure, but we construct the map offline. Unlike Hydra's oversegmented room segmentation, we accurately segment areas. While Hydra focuses on fine-grained indoor object mapping, we target large indoor and outdoor environments, including multi-floor scenes with adaptable layering.

Another similar research direction is parsing the layout of buildings from 2D or 3D data. A considerable amount of work has been dedicated to parsing 2D maps \cite{bormann2016room}, including hypothesis-based methods \cite{kleiner2017solution} and learning-based approaches \cite{liu2018floornet}. One common method for constructing topological maps from 2D occupancy grid maps is to compute Voronoi diagrams, which facilitate the extraction of structures such as points and regions as topological nodes. Depending on the extracted structures, related research can be roughly divided into three types: selecting key points of the Voronoi diagram, such as region boundaries (e.g., doors), to divide the environment into disjoint regions \cite{wurm2008coordinated, thrun1998learning}; using the vertices of the Voronoi diagram as nodes in the topological map \cite{schwertfeger2016map, setalaphruk2003robot}; performing region segmentation based on the Voronoi diagram, treating regions as nodes, as demonstrated in the work by Friedman et al. \cite{Friedman2007VoronoiRF} using Conditional Random Fields for labeling. 

In addition to Voronoi-based methods, Bormann et al. \cite{bormann2016room} introduced several approaches to parse 2D maps into region nodes such as rooms, including morphology-based segmentation, distance transform-based methods, and feature-based segmentation.
Hou et al. \cite{Hou2019AreaGraph} proposed a method based on Voronoi diagrams to construct a topological graph with regions as nodes and corridors as edges, using a non-heuristic region growing approach to identify meaningful areas such as rooms. This method is used as the basis for the automatic generation of the topological graph in this paper. Recent work has focused on 3D data. Liu et al. \cite{liu2018floornet} and Stekovic et al. \cite{stekovic2021montefloor} projected 3D point clouds onto 2D maps, but this approach is not suitable for multi-story buildings. Zheng et al. \cite{zheng2021research} detected rooms by performing region growing on a 3D metric-semantic model.

\section{osmAG}
In this section, we will provide a detailed exposition of the structure and composition of the osmAG file format. Our format utilizes the open street map XML tag ``way" for areas and passages of the Area Graph, standardizes the use of some common attributes for the XML way  elements and introduces a couple of osm tags. osmAG maps can be embedded into normal osm maps - since osmAG is fully compatible with osm.

\subsection{Definition}

\begin{figure}[tb]
	\centering
	\includegraphics[width=0.95\linewidth]{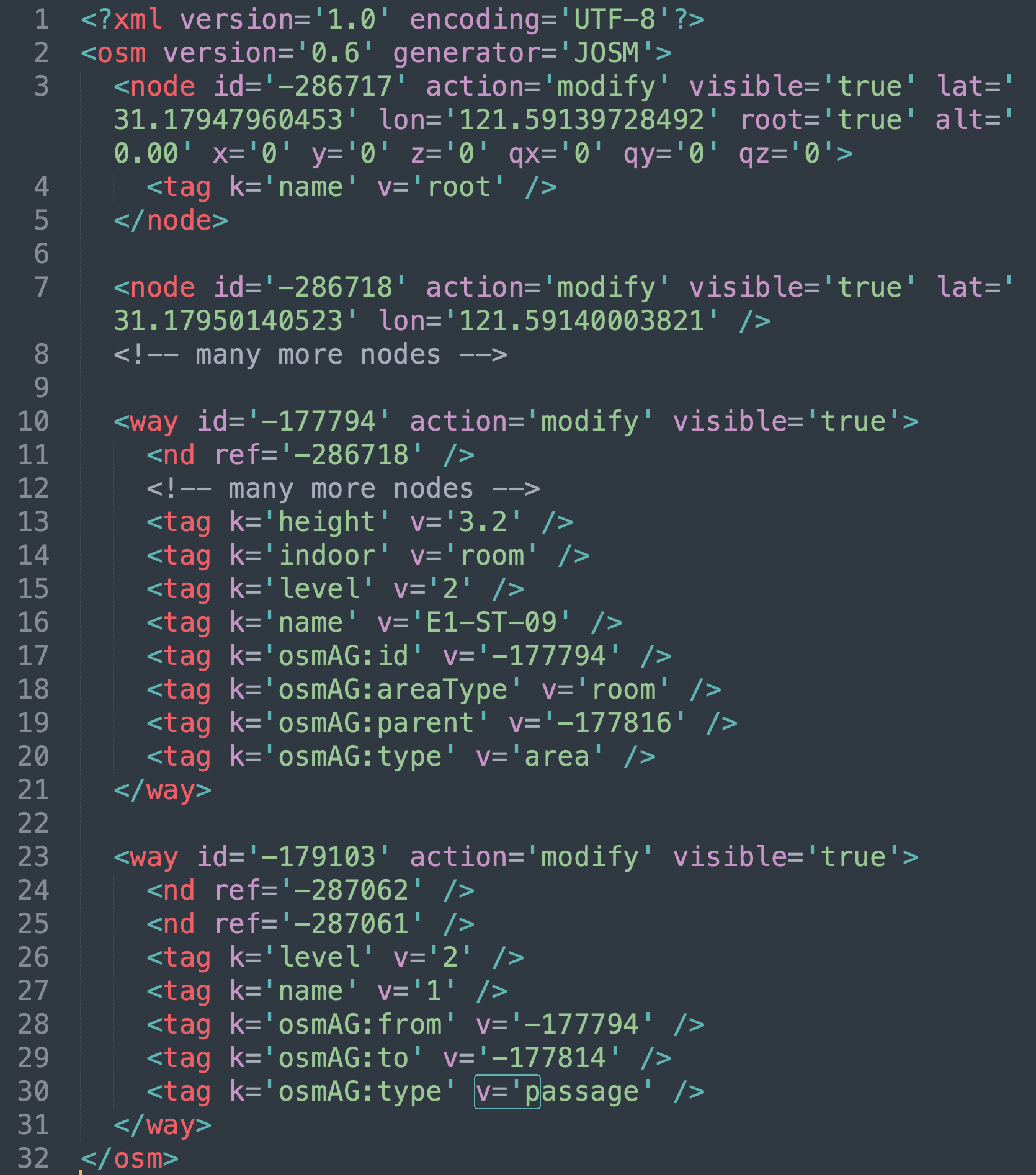}
	\caption{Non-functional demo of the osm xml of osmAG}
	\label{fig:xml}
\end{figure}

First we define some basic osmAG elements, followed by their on-disk osm XML storage format (see Fig. \ref{fig:xml}). In the beginning we just consider a non-hierarchical map - the more complex definition for the hierarchy follows afterwards.



\vspace{5pt}
\noindent\textbf{Area: }
The node of an AreaGraph that encodes an area as a closed polygon, where the points are stored in an ordered list of nodes. Often, especially indoors, areas represent rooms or corridors. The construction algorithm has a parameter that decides how to segment the space (minimum free distance). But areas can also represent whole buildings, certain types of streets or terrains or a whole campus. 

\vspace{5pt}
\noindent\textbf{Inner area: }
A free area surrounded by a boundary, e.g. a room, such as the blue outline in Fig. \ref{fig:localization}. This information is useful for example for localization, where you want to know if you should match against this wall from the inside (inner area) or the outside.


\vspace{5pt}
\noindent\textbf{Structure area: }
The space containing a closed boundary, which is the outer contour of the area and usually is a parent of multiple sub-areas, for example the outer walls of a building. 

\vspace{5pt}
\noindent\textbf{Passage: }
The edge of the Area Graph, connecting two areas topologically.  The two areas have to share at least two points (nodes). Those shared points form the passage, which is represented as a set of points forming a line.

\vspace{5pt}
The on-file format of osmAG follows the open street map (osm) standard with some extensions:

\noindent\textbf{XML tag: node: }
A geographic location as latitude and longitude coordinates with an id. 

\vspace{5pt}
\noindent\textbf{Root node: }
A node tag that contains the pose of the map origin. It converts the WGS84 geodetic coordinate system to Cartesian coordinates as a reference point and facilitates traditional robot indoor navigation. 

\vspace{5pt}
\noindent\textbf{XML tag: way: }
An ordered lists of nodes - the osm method to encode both streets and areas.


\vspace{5pt}
\noindent\textbf{osm way key: osmAG:id: }
While osm is unaware of our tags, we cannot utilize the osm ids for ways, as they may change when uploading the data to an osm server. Therefore, we create our own ids for them.

\vspace{5pt}
\noindent\textbf{osm way key: osmAG:type: }
If present that osm way is part of the Area Graph. The value encodes whether this way is an area or a passage.

\vspace{5pt}
\noindent\textbf{osm way key: osmAG:areatype: }
Inner or structure, see above.

\vspace{5pt}
\noindent\textbf{osm way keys: osmAG:from  and osmAG:to: }
While the Area Graph is undirected, for the on-file format we save the two osmAG:ids of the two areas connected to a passage in the values of the osmAG:from and osmAG:to keys in arbitrary order. 

\vspace{5pt}
\noindent\textbf{osm way key: osmAG:parent: }
This is the key that encodes the hierarchical structure. Its value is the osmAG:id of this area's parent. The areas of all children have to be fully contained within their parent. Areas can never overlap, unless they are on different elevations or are parent and (grand-)child. 

This spans a hierarchical tree of areas and their children. One osmAG file can contain multiple such trees, together with traditional osm vector data. 

The hierarchical structure is mainly used for two scenarios: First, we may want to join multiple neighboring areas into one bigger area, e.g. to facilitate faster planning and encode bigger areas and semantic tags for them. For example, we can, in the outdoor case, join areas for a parking place and neighboring areas for streets together to a bigger area. Indoors we can join multiple rooms to form a ``floor" of a building.

In the second case, the hierarchical structure is used to stack several 2D areas on top of each other, e.g. multiple floors of a building. The height of the floors is encoded, following the osm standard, as height above the ground floor in the osm key ``height". 

The different levels of a building are connected via passages. E.g. an elevator will have areas on every floor that are connected to each other vertically by special elevator passages and that are also connected in the 2D plane to one (or more) neighboring areas with passage(s) that have the semantic information ``elevatordoor". Stairs follow the same principle. More details can be found on GitHub. 

\subsection{Visualization and Editing}

\begin{figure}[t]
	\centering
	\subfigure[2D view of two osmAG buildings in an osm environment]{
		\centering
		\includegraphics[width=0.22\textwidth]{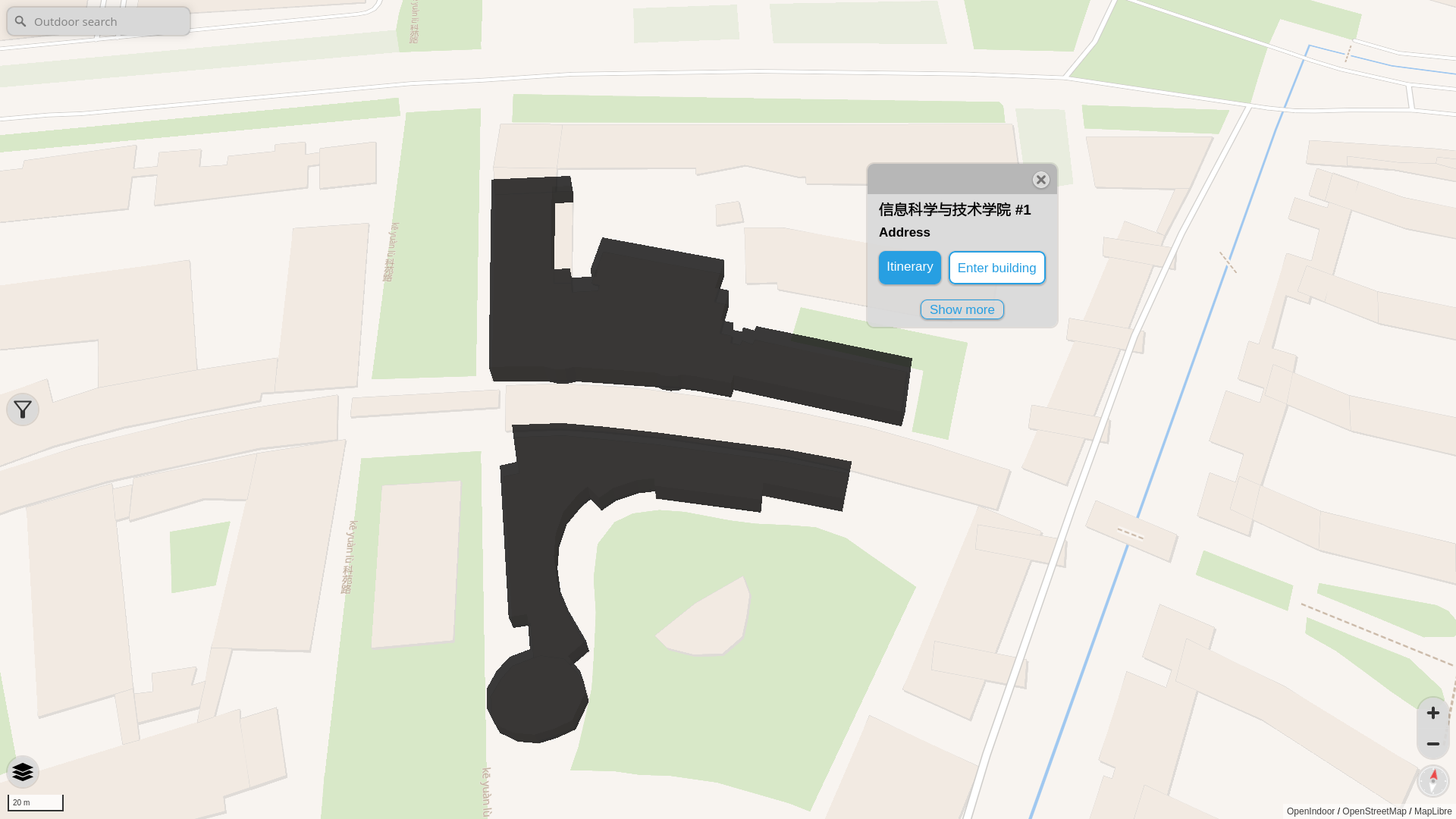}
		\label{a}}
	\centering
	\subfigure[3D view of two floors of one building]{
		\centering
		\includegraphics[width=0.22\textwidth]{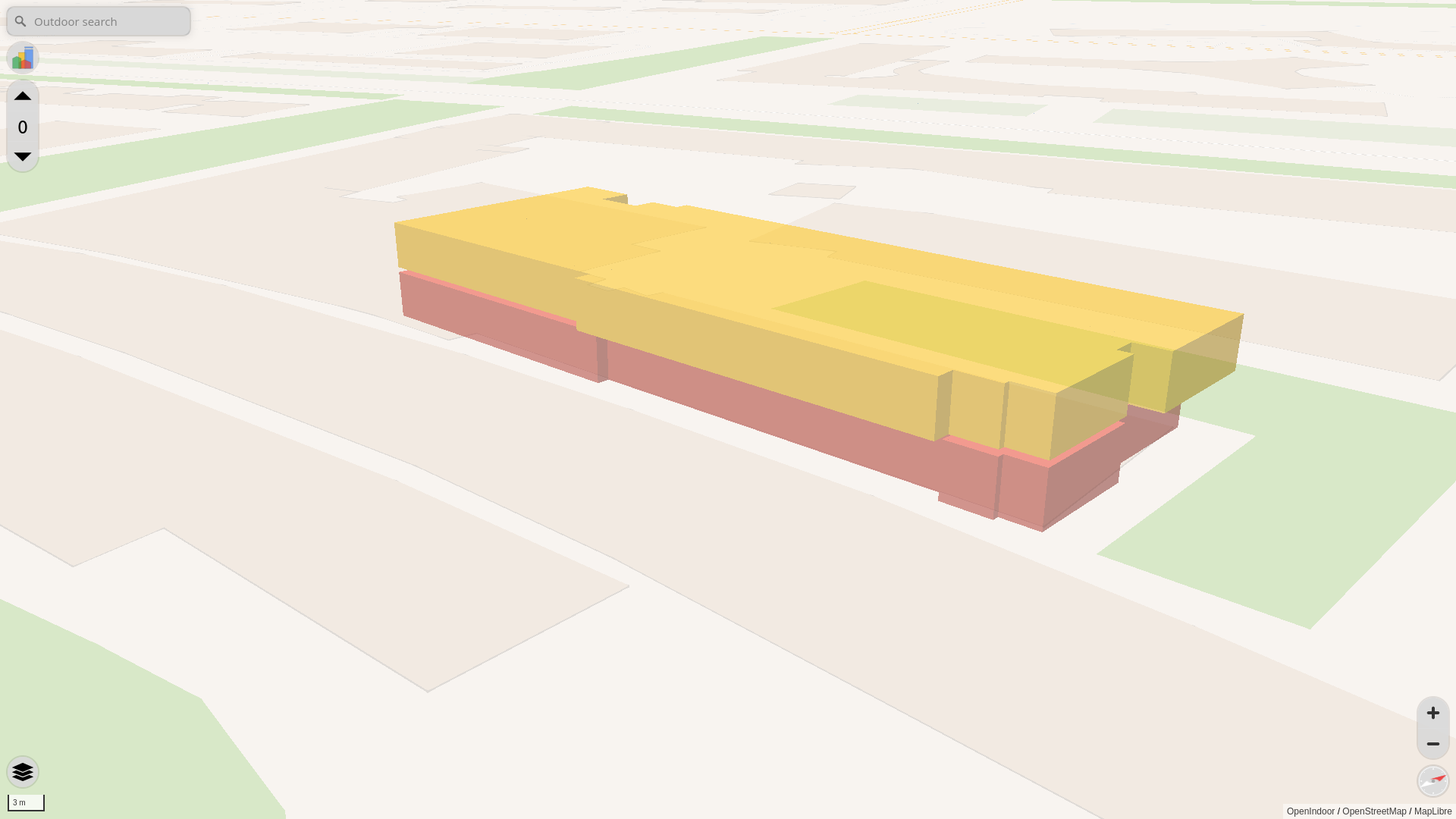}
		\label{b}}
	
	\subfigure[3D view of the first floor]{
		\centering
		\includegraphics[width=0.22\textwidth]{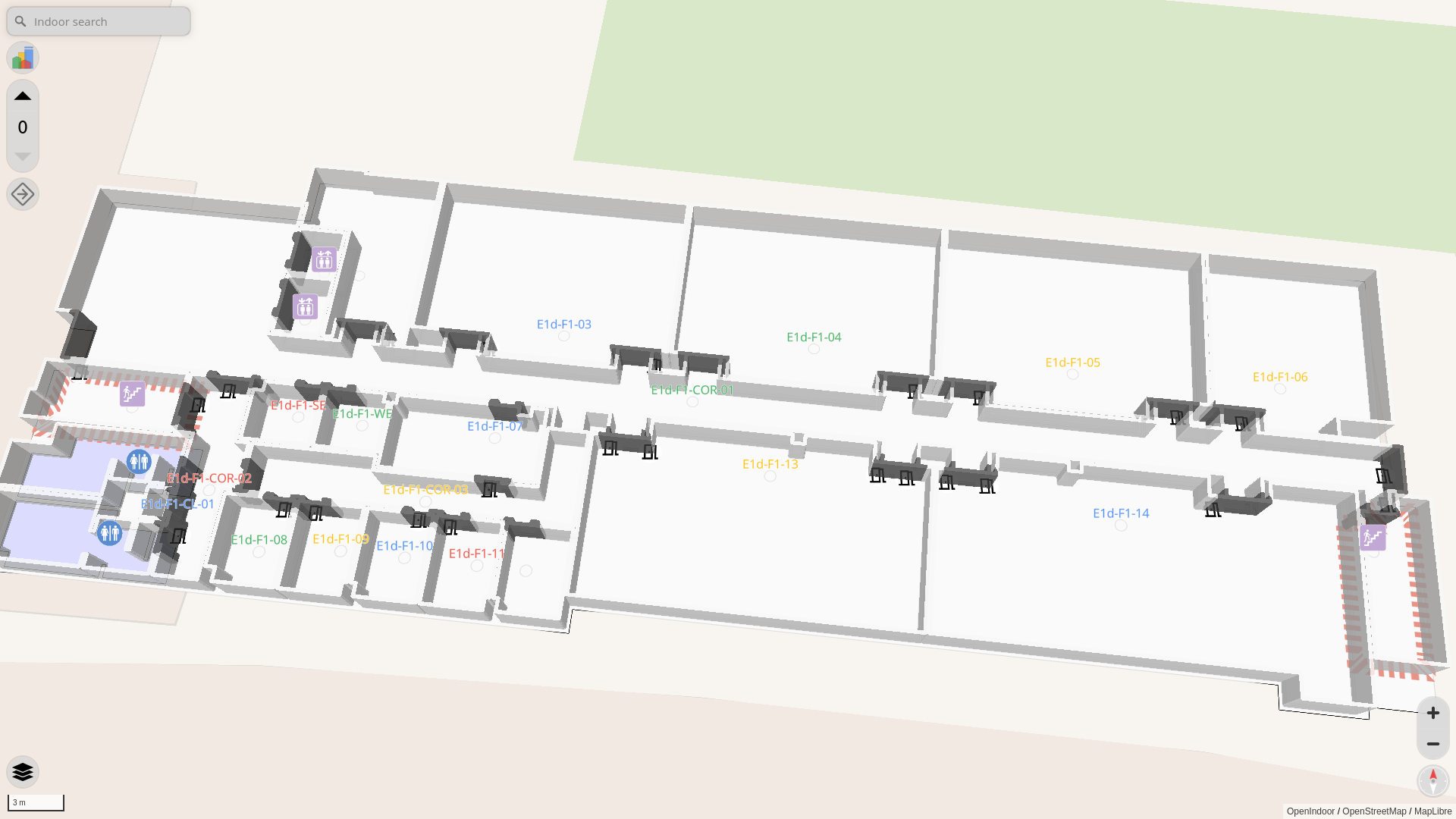}
		\label{c}}
	\subfigure[3D view of the second floor]{
		\centering
		\includegraphics[width=0.22\textwidth]{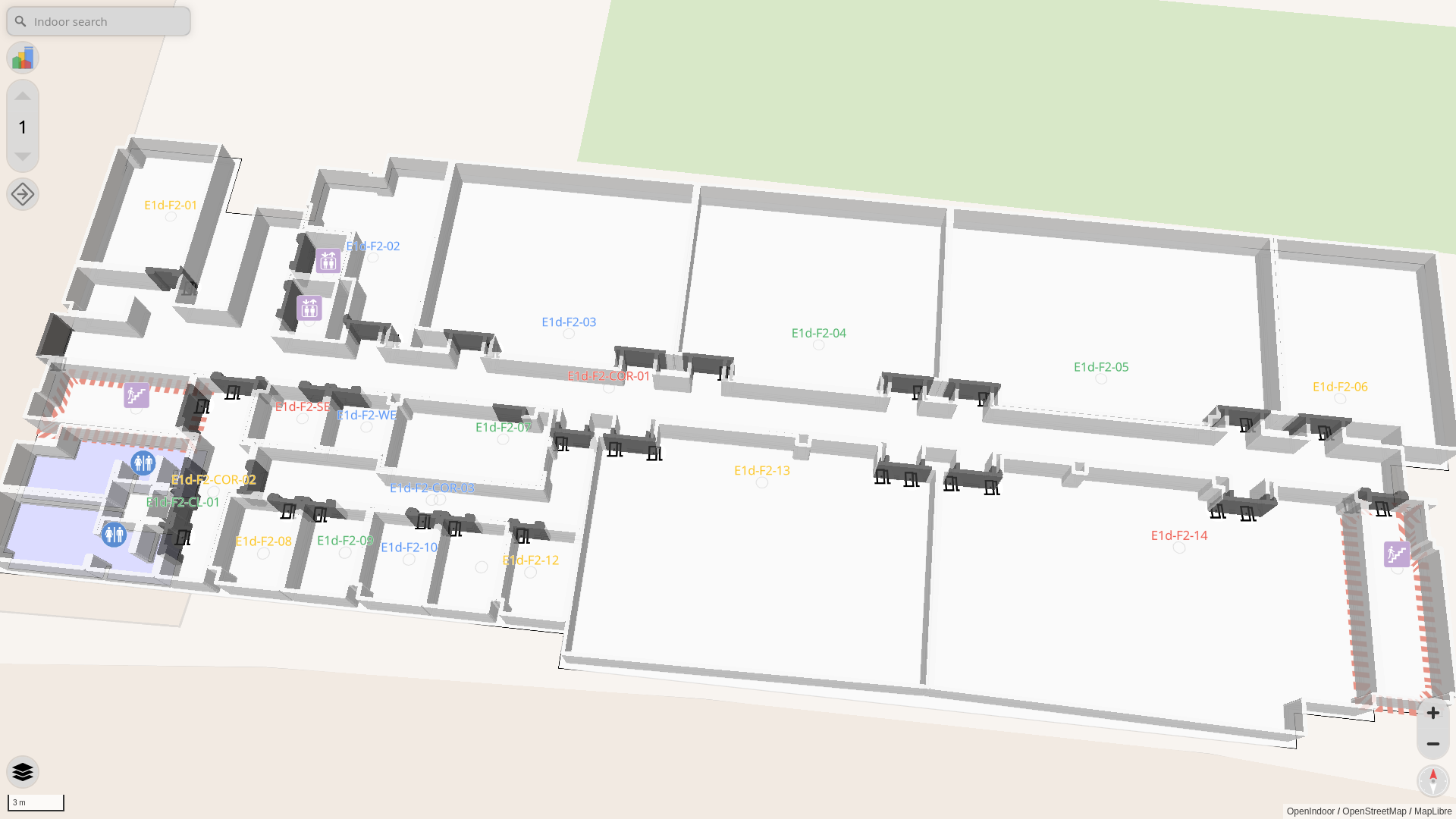}
		\label{d}}
	\caption{3D Rendering of an osmAG map}
	\label{figurelabe4}
\end{figure}

Since osmAG follows the basic data format of open street map, users can perform visualization and editing operations in any software or application that supports OpenStreetMap. The three-dimensional rendering of osmAG data was performed in the OpenIndoor application\footnote{\url{https://wiki.openstreetmap.org/wiki/OpenIndoor}}. Fig. \ref{figurelabe4} (a) illustrates the shape and geographic location of the indoor building. Fig. \ref{figurelabe4} (b) showcases some floors within one of the buildings, highlighting variations in the structural layout between floors. Fig. \ref{figurelabe4} (c) and Fig. \ref{figurelabe4} (d) depict the rendered floor plans of different levels, with black sections representing passages. Walls with a fixed thickness show the area polygons, and semantic labels such as ``stairs", ``restrooms", and ``elevators" are annotated according to the icons used in OpenStreetMap.

\begin{figure}[thpb]
        \centering
        \includegraphics[width=0.45\textwidth]{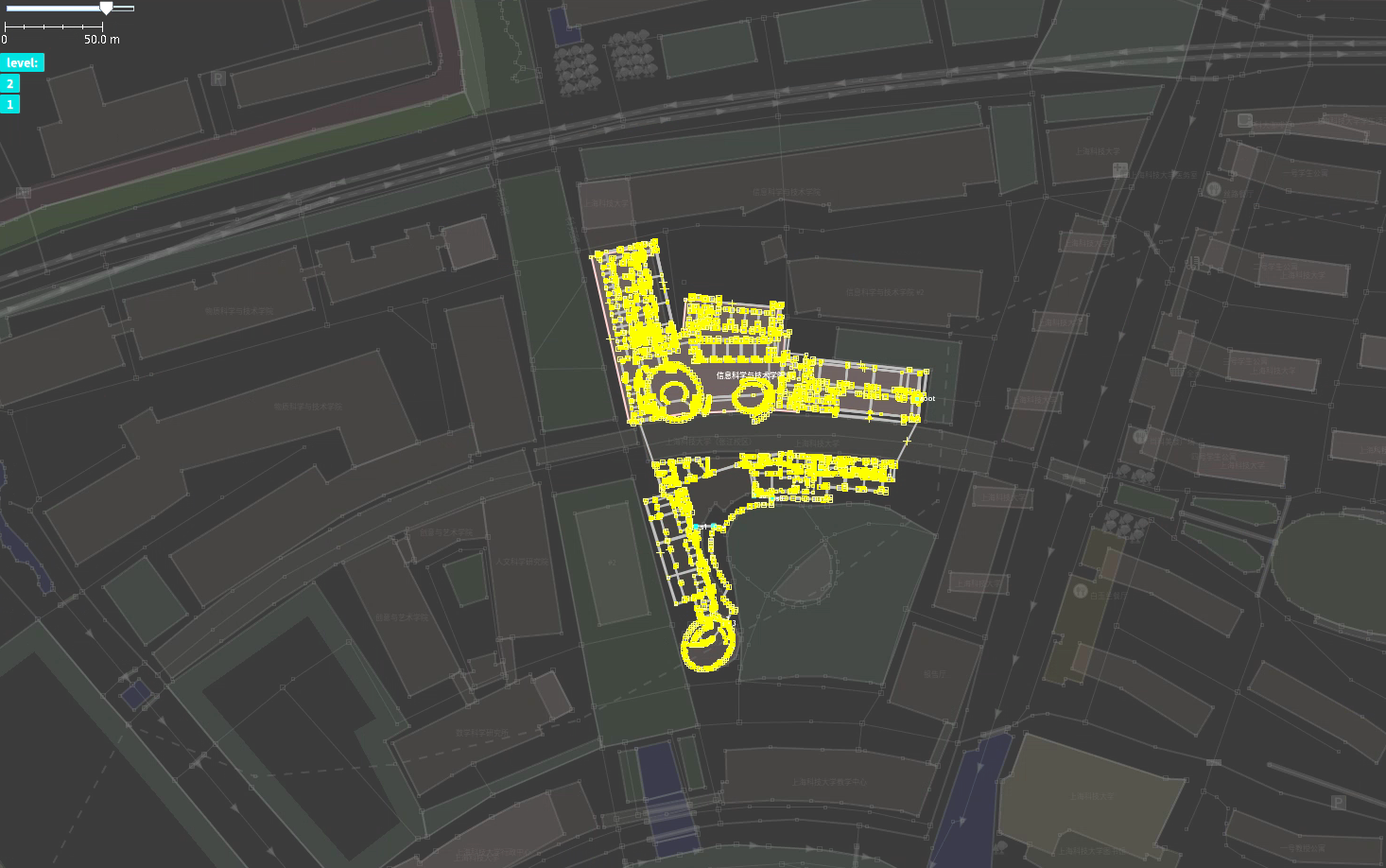}
        \caption{2D visualization of osmAG (highlighted) of two buildings in JOSM, together with classic osm data}
        \label{figurelabe5}
    \end{figure}

The two-dimensional visualization and manual editing of osmAG are achieved through the JOSM (Java OpenStreetMap Editor) application. By using JOSM, users are able to easily view, edit, and update the area graph, as shown in Fig. \ref{figurelabe5}.

\subsection{Automated Generation}

Although existing editors like JOSM can be used to edit area map, in this case, the elements, information, and relationships of the topological map must be manually added. It is impractical when the scene is vast. We focus on the basic functionality of offline mapping for osm. This functionality is implemented by a C++ library based on osmAG.

Offline mapping involves segmenting, extracting, and merging from existing maps to create a new map. These maps can be CAD files of buildings, 3D point clouds, or maps generated by robots. We implement the automatic conversion from 2D grid maps and 3D point cloud maps into osmAG, as well as the simple fusion of multiple topological maps.


\vspace{5pt}
\noindent\textbf{Generating from a 2D grid map.}
Our previous work \cite{Hou2019AreaGraph, hou2022matching} generates non-hierarchical areas graphs from 2D grid maps. We modified that work to output osmAG files that can then be further edited.

    \begin{figure}[t]
        \centering
        \subfigure[A CAD file]{
            \centering
            \includegraphics[width=0.22\textwidth]{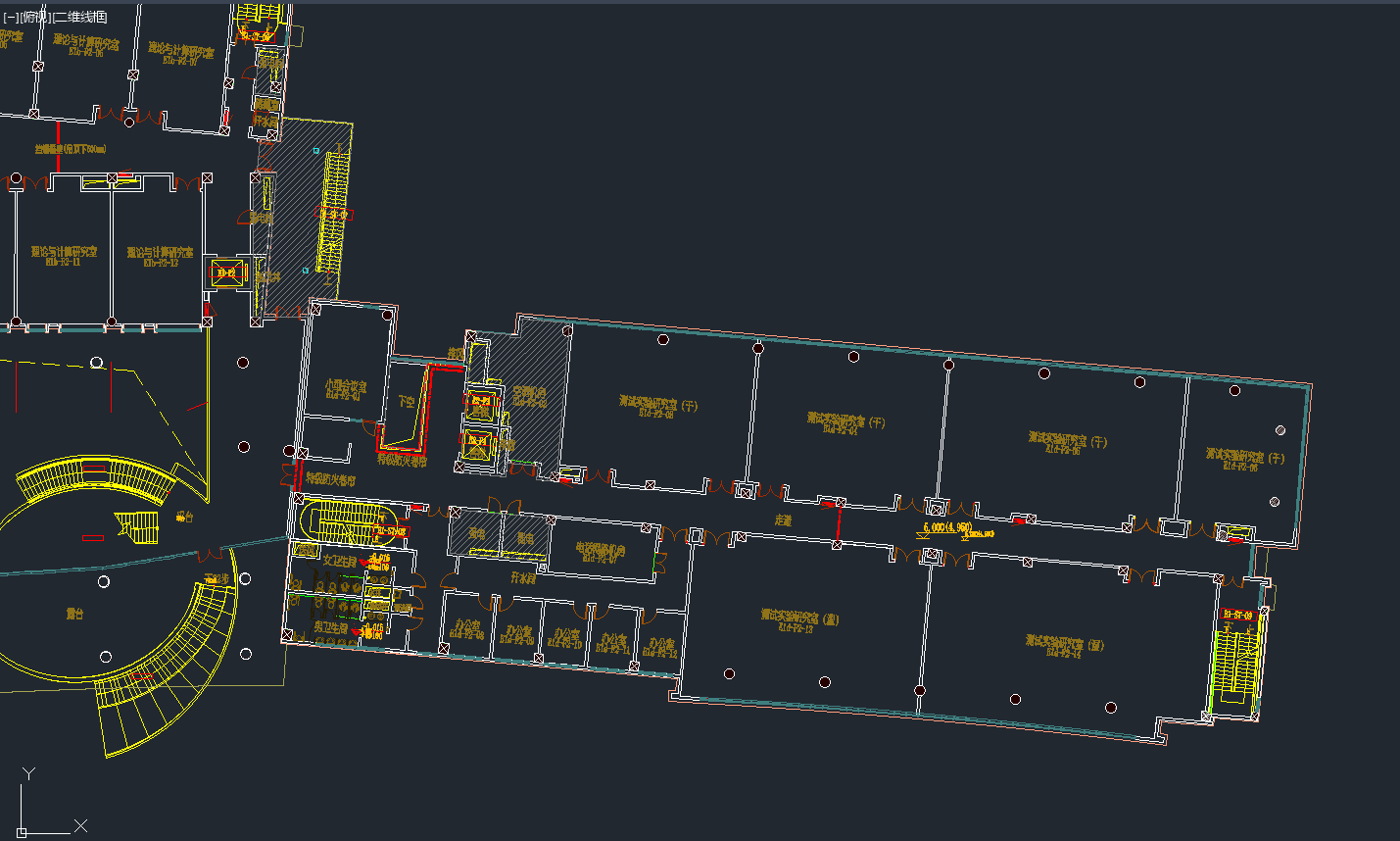}
            \label{a}}
            \centering
        \subfigure[The area graph of that CAD file]{
            \centering
            \includegraphics[width=0.22\textwidth]{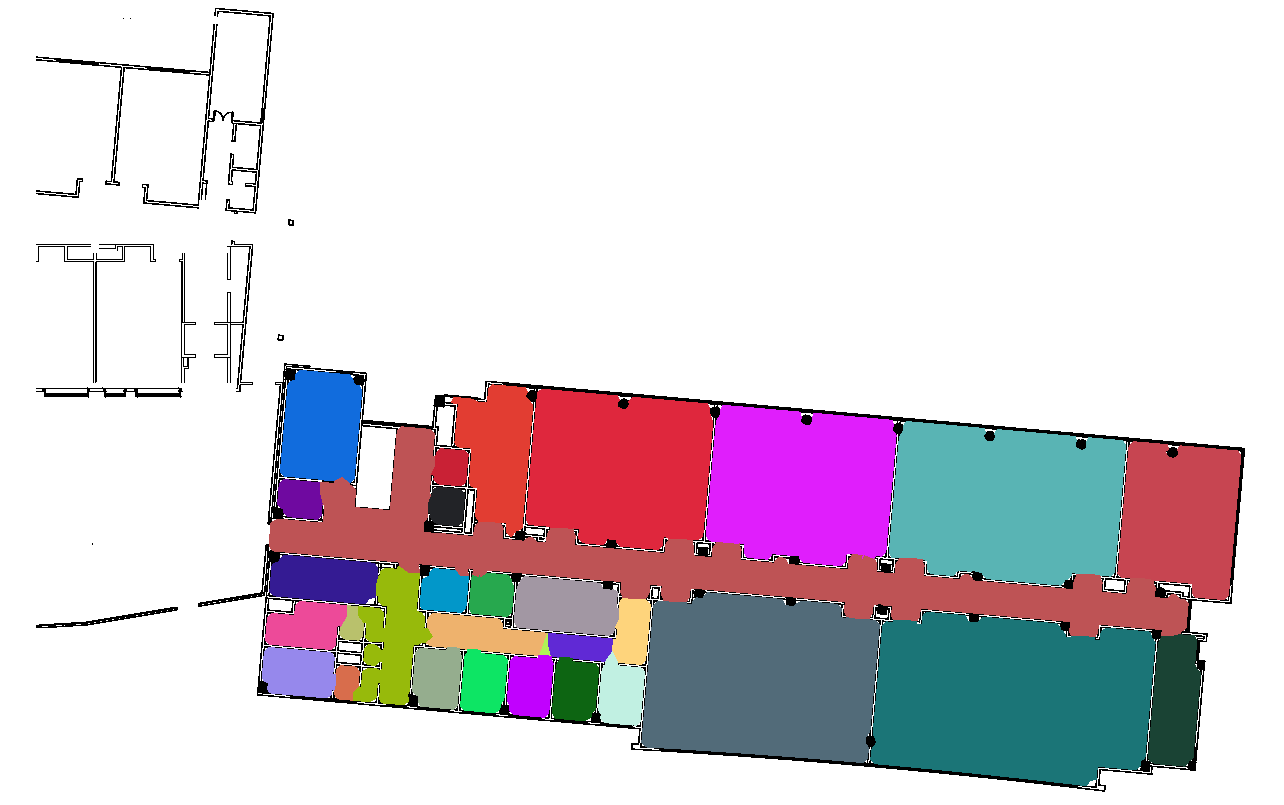}
            \label{b}}

        \subfigure[A draft osmAG visualization]{
            \centering
            \includegraphics[width=0.22\textwidth]{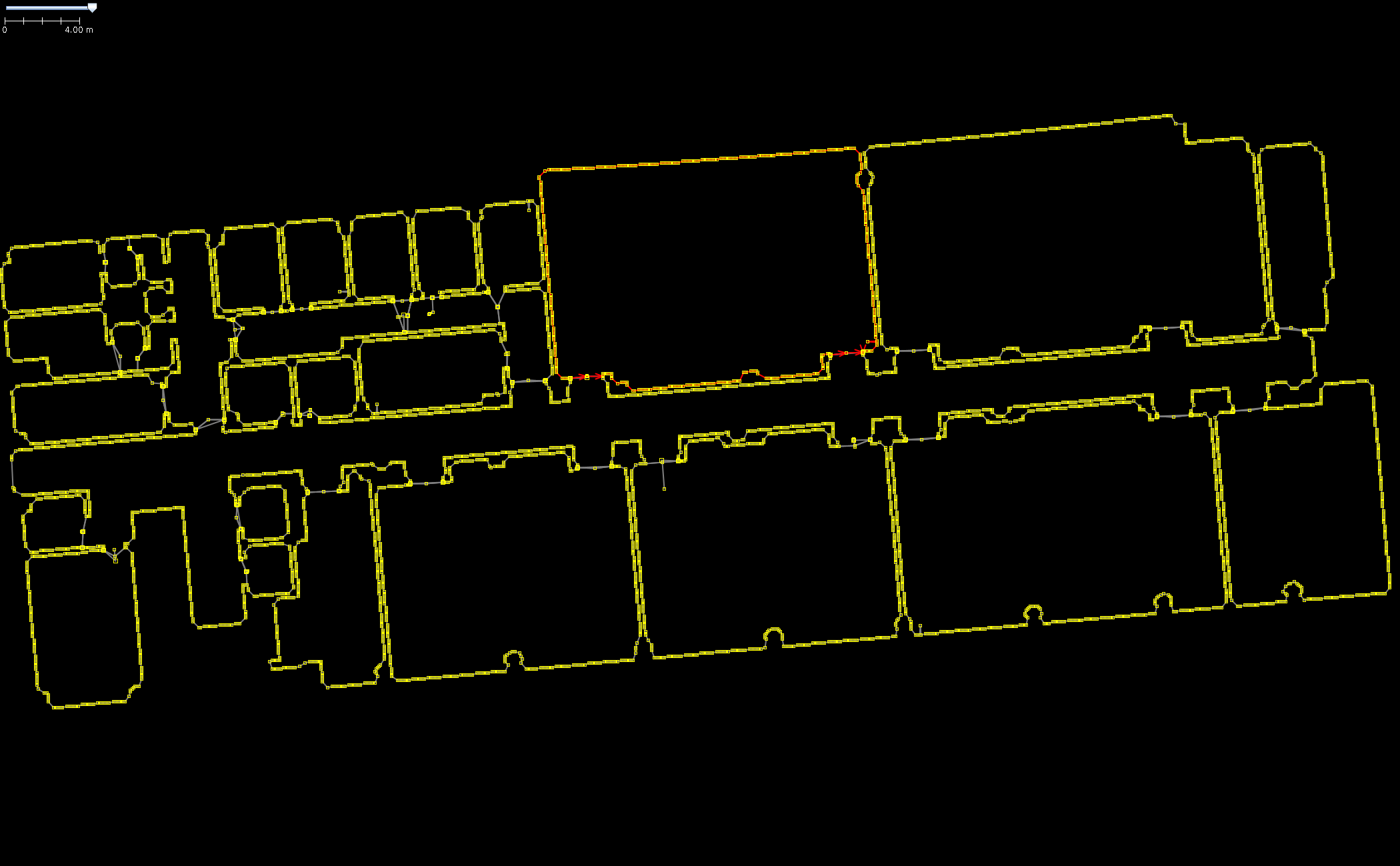}
            \label{c}}
        \subfigure[osmAG with a marked passage]{
            \centering
            \includegraphics[width=0.22\textwidth]{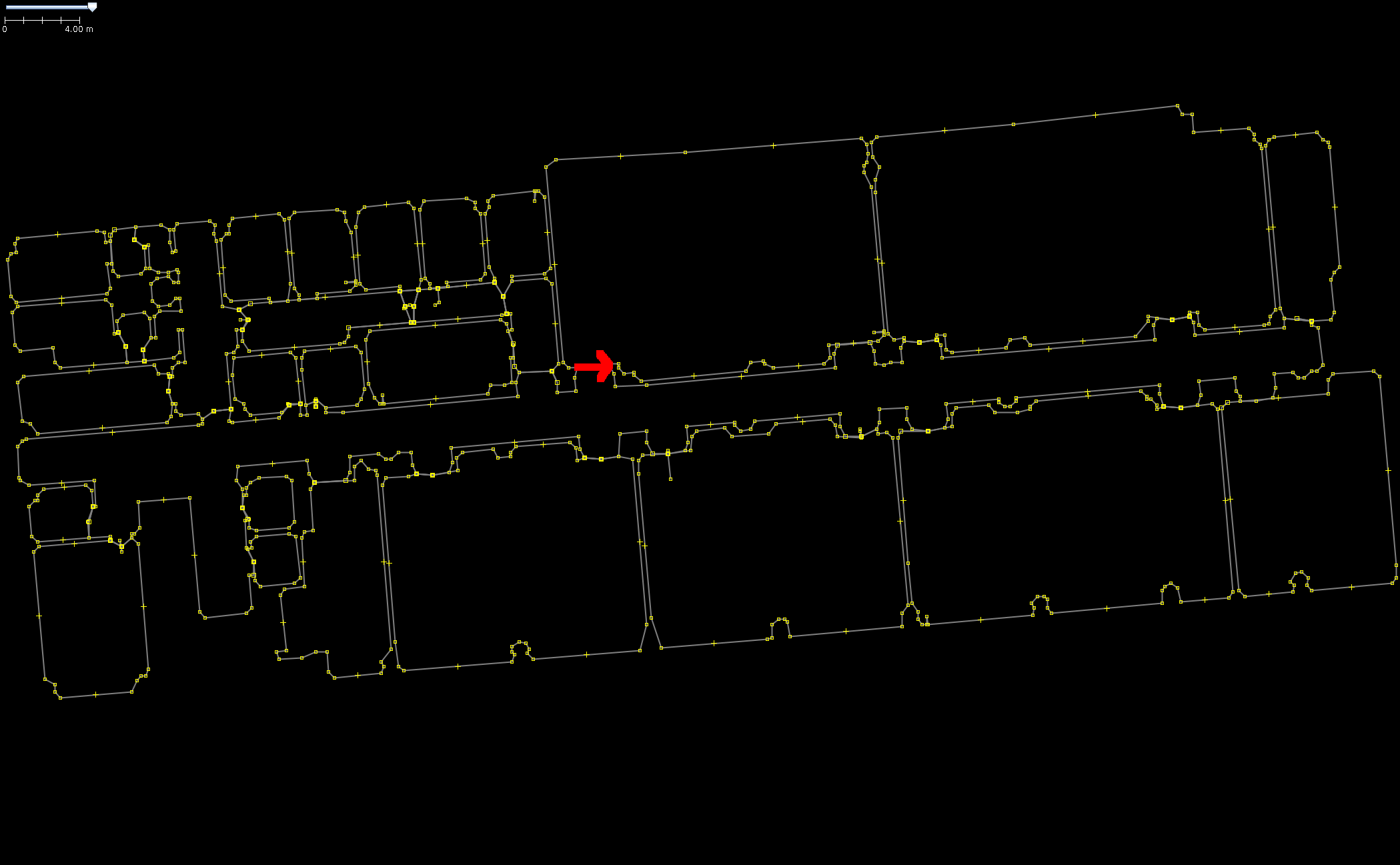}
            \label{d}}
        \caption{Generating osmAG from CAD data}
        \label{figurelabe6}
    \end{figure}

\vspace{5pt}
\noindent\textbf{Generating from a CAD file.}
The process is illustrated in Fig. \ref{figurelabe6} (a), taking a CAD file input as an example. Generally, CAD files for building layouts contain multiple layers and semantic information. Currently, CAD files are processed manually by removing layers. Fig. \ref{figurelabe6} (b) shows the generated area graph, where it can be observed that a corridor in the bottom left of the current image is divided into several areas. By adjusting the parameter $alpha$ shape in the algorithm, issues of over-segmentation and under-segmentation can be alleviated. Fig. \ref{figurelabe6} (c) demonstrates the effect of directly using the Area Graph method to generate the topological map's area nodes. Fig. \ref{figurelabe6} (d) illustrates how this method correctly generates inner areas and passages for the osmAG (the red arrow in the image represents a passage).

\vspace{5pt}
\noindent\textbf{Generating from a 3D pointcloud map.}
We follow our previous work \cite{he2021hierarchical} which generates hierarchical 3D topometric maps from 3D point clouds of the whole building, to extract 2D Area Graphs from a 3D pointcloud map.




    \begin{figure}[thpb]
        \centering
        \subfigure[One floor automatically extracted from the 3D building point cloud]{
            \centering
            \includegraphics[width=0.45\textwidth]{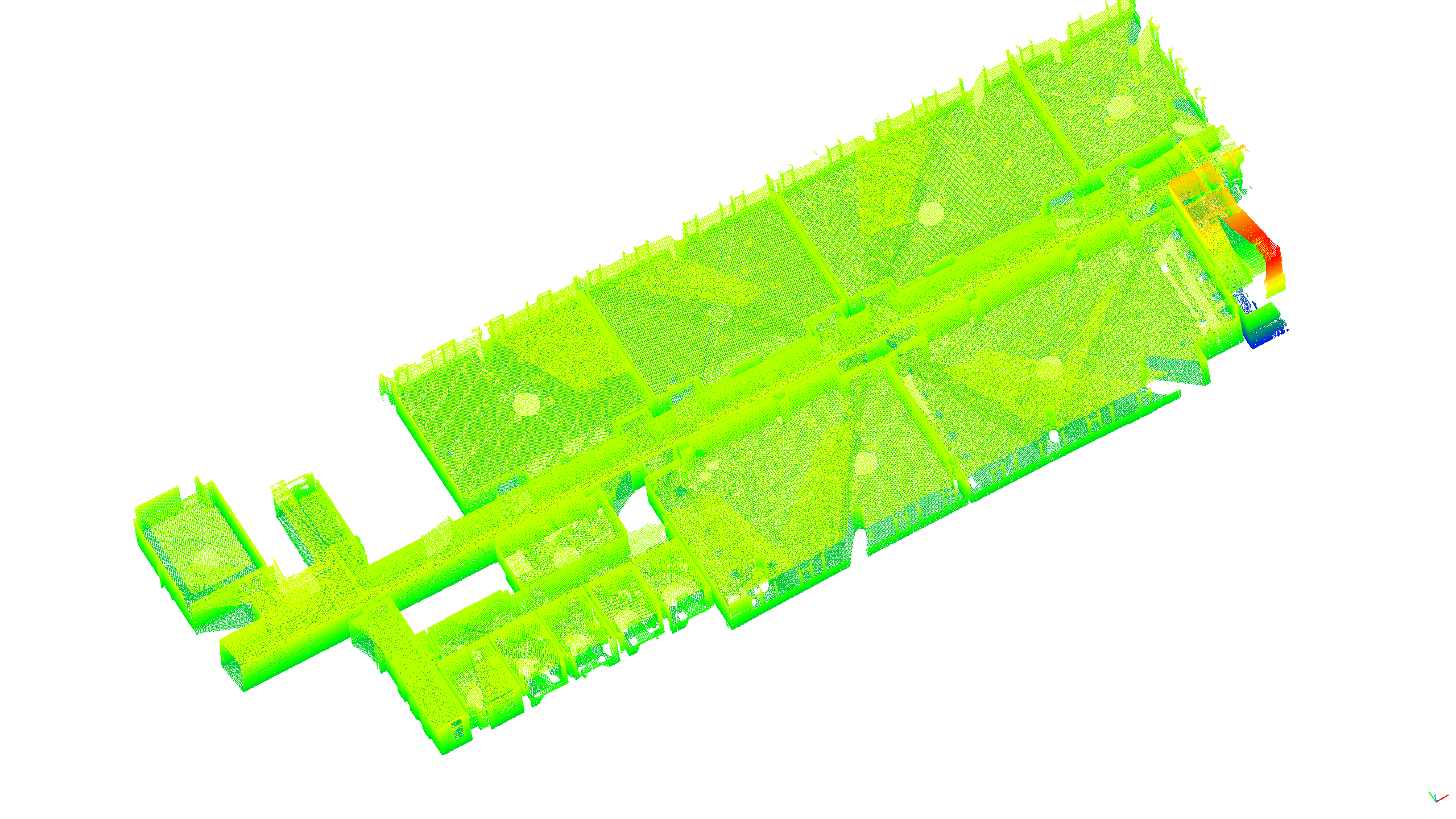}
            \label{a}}

        \subfigure[2D Area Graph extracted from the 3D point cloud]{
            \centering
            \includegraphics[width=0.45\textwidth]{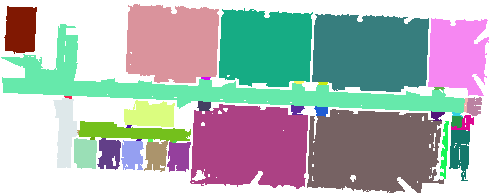}
            \label{b}}
        \caption{The automatic generation of osmAG from 3D maps\cite{he2021hierarchical}}
        \label{fig:from3D}
    \end{figure}


These extraction methods are suitable for generating topological representations in indoor environments.
For further processing, our osmAGlib C++ library efficiently merges multiple maps. This is achieved through a process that involves assessing the distances between nodes and selectively consolidating certain points to merge osmAG maps and integrate existing outdoor OpenStreetMaps into the current~map. 

\section{Path Planning with osmAG}

Navigation is an essential part of mobile robotics and utilizing osmAG for path planning is a crucial part in developing a navigation stack for osmAG. We have already demonstrated the effectiveness of path planning in the area graph \cite{hou2018topological}. Here, we describe our approach to planning with the hierarchical area graph data structure. 

The goal of path planning in osmAG is to provide the best global path between two coordinates. The paths generated may be close to walls and it is the job of local planning to keep a safe distance during navigation. 

\begin{figure}[tb]
	\centering
	\subfigure[Planning through the center of an area]{
		\centering
		\includegraphics[width=0.28\textwidth]{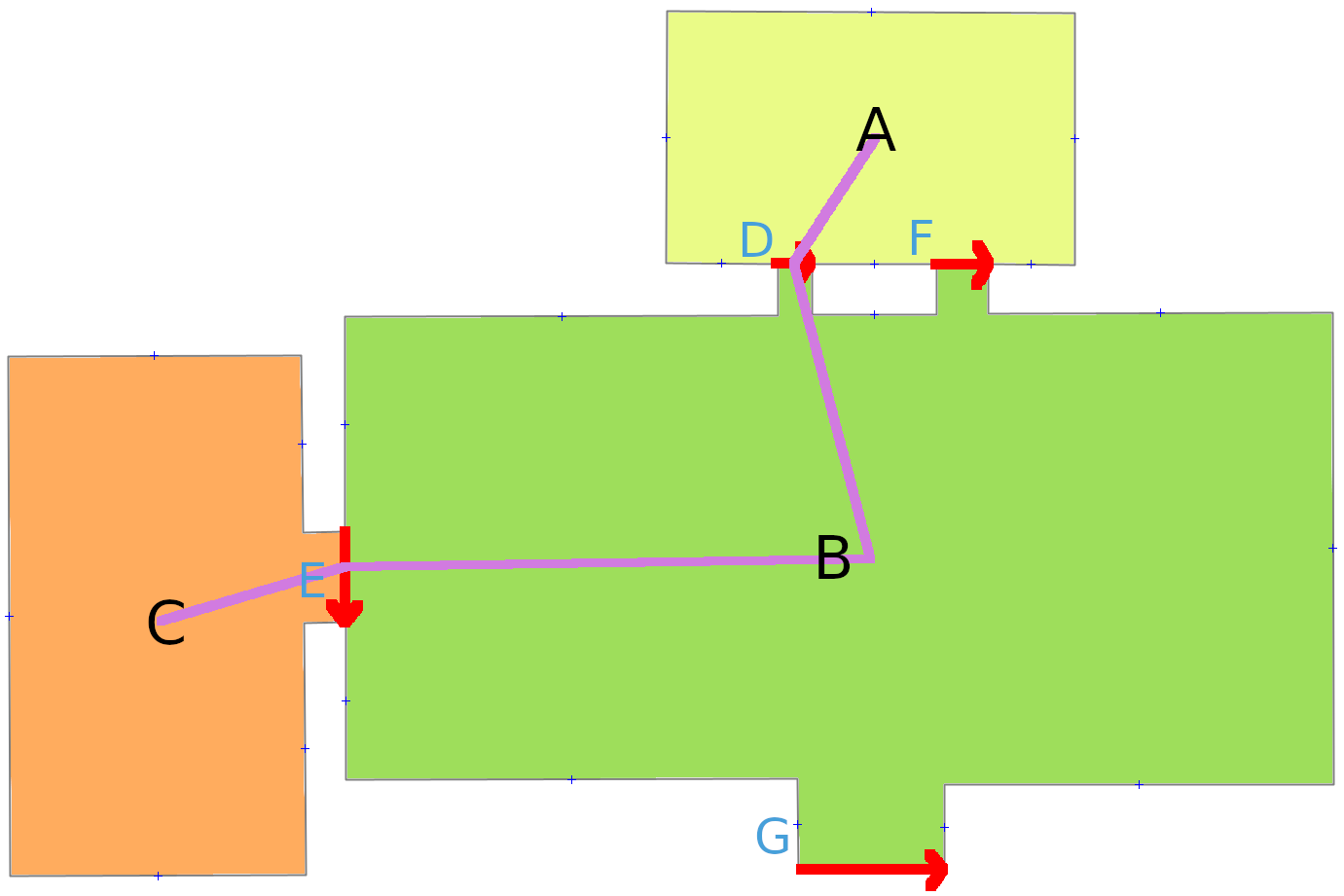}
		\label{a}}
	
	\subfigure[Planning true cost with rendered 2D grid map]{
		\centering
		\includegraphics[width=0.28\textwidth]{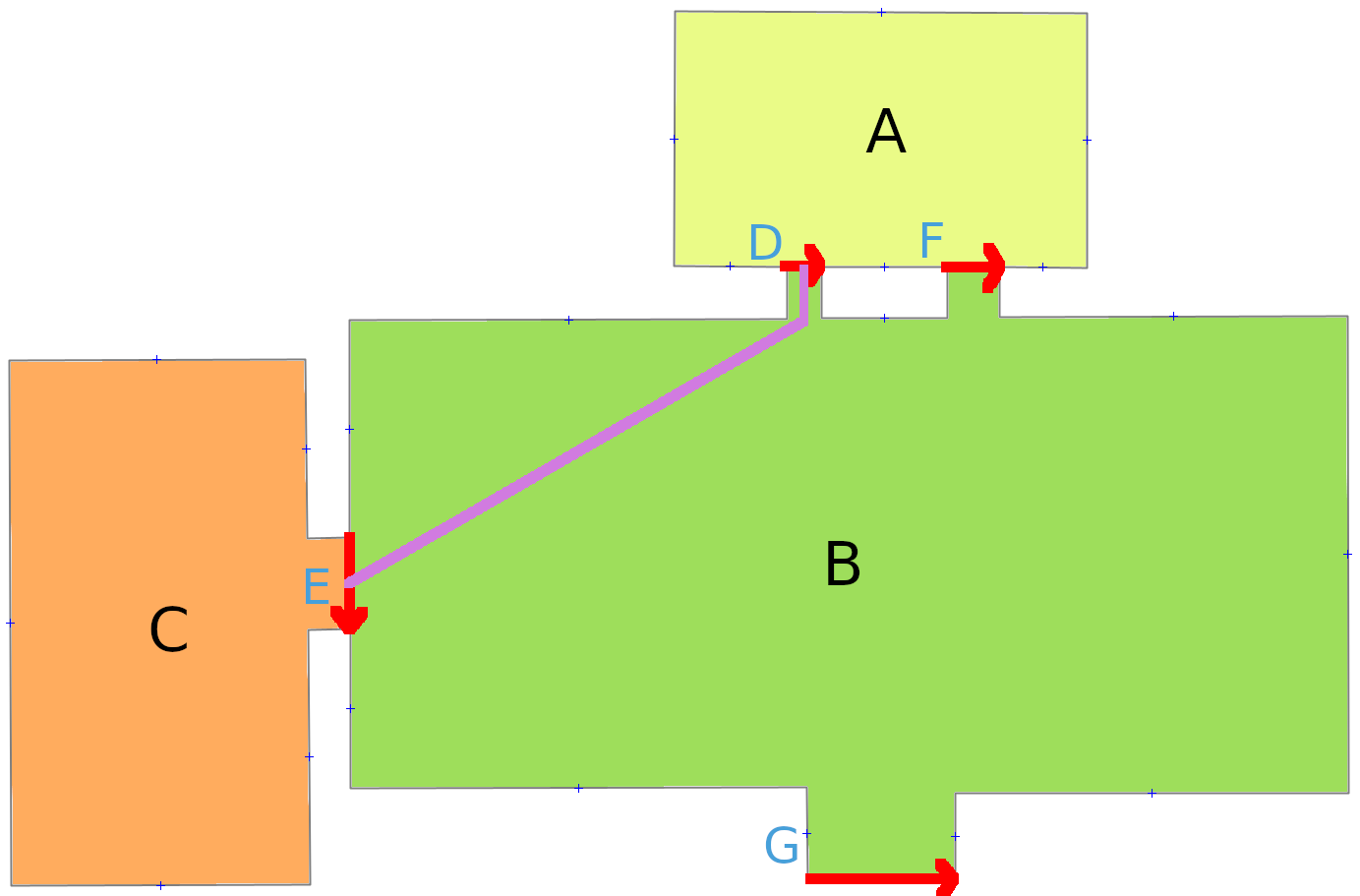}
		\label{b}}
	\caption{Comparison of cost calculation options during path planning.}
	\label{figurelabe8}
\end{figure}

We cannot use the graph structure of the area graph to directly plan, because the cost is in traversing the vertices, while the cost of traversing a passage is typically zero. The cost of traversing the area depends on which passage you enter and through which passage you exit. A simple solution would assign the cost by summing up the distance from the first passage to the center with the distance from the center to the second passage. But that may overestimate the cost and lead to sub-optimal paths as shown in Fig. \ref{figurelabe8} (a). 

We therefore opt for generating a 2D grid map just from that area and utilizing A* on that grid map to calculate the true cost for that traversal (Fig. \ref{figurelabe8} (b)). Consequently, for planning we build a new graph where the passages are the vertices and where we add edges to all possible other passages of that area.

We pre-compute this passage graph on the leaves of the whole Area graph. Planning a path between two coordinates then follows these simple steps:
\begin{enumerate}
	\item Find the leaf areas of the start and goal points.
	\item (Special case: if both are in the same area, render that area and use A* on that grid map)
	\item Temporarily add edges from the start (goal) point to all passages of that area by using A* on the rendered 2D grid map of that area.
	\item In that graph use A* to search for a path from the start to the goal node.
	\item Return the path in terms of a list of passages and areas and/or in terms of a set of points (from the A* of the pre-computed 2D grids). 
\end{enumerate}


\subsection{Speedup using Hierarchy}

Path planning can be further sped up by utilizing bigger, higher-level areas. This is done by pre-computing the cost of all combinations of start and end passages of each non-leaf area. This pre-computation is very fast, since it utilizes the already computed costs of the child areas. During planning, whenever we encounter a passage we select the highest level area it connects to that is not containing the goal point - unless it is the goal leaf area. 

\subsection{Planning using Semantic Information}
One great feature of planning with the osmAG map is, that it can take robot-specific capabilities into account. For that semantic cues from the passages and areas are taken into account. For example, a wheeled robot will not plan through an area with the semantic information ``stairs", while a legged robot might. For that we plan to callback a cost-calculation function for every area and passage that is encountered, so robot specific code may determine the cost. The legged robot may return the cost in terms of the estimation of time needed to traverse and still avoid the stairs---if there is a ramp with just a small detour close by. Other information taken into account may be the step height of a curb (bigger wheeled robots can traverse, smaller ones cannot), the type of door (automatic, push, pull),  or the material of an area (pavement, grass, sand).

\section{Experiment}
Our experiment verifies the feasibility and effectiveness of the proposed global path planning method in osmAG.


This global path planning experiment is conducted in an osmAG that includes two multi-story buildings and an outdoor area. The osmAG is stored in XML file format with a size of 883.9 kb. It consists of 4908 nodes, 500 areas, and 347 passages. The actual area of the indoor area is approximately 22300 $m^2$, while the outdoor area is approximately 3000 $m^2$. Fig. \ref{figurelabe0} shows how we use the proposed osmAG to represent a complex and large-scale environment, and also demonstrates the global path planning results of the robot spanning two floors, across the road, leveraging semantic cues such as whether to use elevators.
The path shown by the black line in the figure starts from the first floor of the building located in the bottom half of the figure, through the outdoor area to the first floor of the building in the top half of the figure, and then through the elevator to the destination located on the second floor. Our experiments are carried out on an Intel i7-6700 CPU. The distance of the planned path is 157.57m (Includes the vertical distance between floors), the time is about 3100 $\mu$s at the topological level, and the pre-compute time (calculating the metric level distance for all areas and can be applied multiple times) is about 200 ms.

\section{CONCLUSIONS}

This paper proposed a novel hierarchical semantic topological map called open street map Area Graph (osmAG), based on the Area Graph and OpenStreetMap. This map represents the environment using a topometric, hierarchical semantic structure with areas as vertices and passages as edges. It can encode outdoor and indoor maps on multiple levels. Through its use of open street map format, we stay compatible with a plenitude of tools and libraries. Our osmAG C++ library is capable of loading osmAG, and visualizing and planning with it in ROS. We also will publish tools to semi-automatically generate osmAG from 2D grid maps, 3D point clouds, and CAD data. 

Through utilizing osmAG, we are able to plan paths for mobile robots, taking into account their capabilities through semantic cues, and planning through multiple floors. We are able to plan very quickly by utilizing the abstract topometric and hierarchical data structure.  Our experiment shows the good performance of our approach. The code and example data are available on GitHub.

For future work, we plan on providing a full ROS osmAG Navigation stack including localization, planning, and navigation with online changing of the osmAG to facilitate real-time updates of temporary or permanent map changes. We will furthermore introduce WiFi access point mapping to include those locations in the osmAG maps to fully automate our osmAG localization approach. We will then test the whole osmAG Navigation stack on real robots and provide it as open source to the robotics community.










\bibliographystyle{IEEEtran}
\bibliography{IEEEabrv, refs}

\begin{thebibliography}{10}
\providecommand{\url}[1]{#1}
\csname url@samestyle\endcsname
\providecommand{\newblock}{\relax}
\providecommand{\bibinfo}[2]{#2}
\providecommand{\BIBentrySTDinterwordspacing}{\spaceskip=0pt\relax}
\providecommand{\BIBentryALTinterwordstretchfactor}{4}
\providecommand{\BIBentryALTinterwordspacing}{\spaceskip=\fontdimen2\font plus
\BIBentryALTinterwordstretchfactor\fontdimen3\font minus
  \fontdimen4\font\relax}
\providecommand{\BIBforeignlanguage}[2]{{%
\expandafter\ifx\csname l@#1\endcsname\relax
\typeout{** WARNING: IEEEtran.bst: No hyphenation pattern has been}%
\typeout{** loaded for the language `#1'. Using the pattern for}%
\typeout{** the default language instead.}%
\else
\language=\csname l@#1\endcsname
\fi
#2}}
\providecommand{\BIBdecl}{\relax}
\BIBdecl

\bibitem{kostavelis2015semantic}
I.~Kostavelis and A.~Gasteratos, ``Semantic mapping for mobile robotics tasks:
  A survey,'' \emph{Robotics and Autonomous Systems}, vol.~66, pp. 86--103,
  2015.

\bibitem{Thrun2003RoboticMA}
S.~Thrun \emph{et~al.}, ``Robotic mapping: A survey.''\hskip 1em plus 0.5em
  minus 0.4em\relax Citeseer, 2002.

\bibitem{he2021hierarchical}
Z.~He, H.~Sun, J.~Hou, Y.~Ha, and S.~Schwertfeger, ``Hierarchical topometric
  representation of 3d robotic maps,'' \emph{Autonomous Robots}, vol.~45,
  no.~5, pp. 755--771, 2021.

\bibitem{Hou2019AreaGraph}
J.~Hou, Y.~Yuan, and S.~Schwertfeger, ``Area graph: Generation of topological
  maps using the voronoi diagram,'' in \emph{19th International Conference on
  Advanced Robotics (ICAR)}, IEEE Press.\hskip 1em plus 0.5em minus 0.4em\relax
  IEEE Press, 2019.

\bibitem{hou2022matching}
J.~Hou, Y.~Yuan, Z.~He, and S.~Schwertfeger, ``Matching maps based on the area
  graph,'' \emph{Intelligent Service Robotics}, pp. 1--26, 2022.

\bibitem{schwertfeger2016map}
S.~Schwertfeger and A.~Birk, ``Map evaluation using matched topology graphs,''
  \emph{Autonomous Robots}, vol.~40, no.~5, pp. 761--787, 2016.

\bibitem{hentschel2010autonomous}
M.~Hentschel and B.~Wagner, ``Autonomous robot navigation based on
  openstreetmap geodata,'' in \emph{13th International IEEE Conference on
  Intelligent Transportation Systems}.\hskip 1em plus 0.5em minus 0.4em\relax
  IEEE, 2010, pp. 1645--1650.

\bibitem{naik2019semantic}
L.~Naik, S.~Blumenthal, N.~Huebel, H.~Bruyninckx, and E.~Prassler, ``Semantic
  mapping extension for openstreetmap applied to indoor robot navigation,'' in
  \emph{2019 International Conference on Robotics and Automation (ICRA)}.\hskip
  1em plus 0.5em minus 0.4em\relax IEEE, 2019, pp. 3839--3845.

\bibitem{wang2018data}
Z.~Wang and L.~Niu, ``A data model for using openstreetmap to integrate indoor
  and outdoor route planning,'' \emph{Sensors}, vol.~18, no.~7, p. 2100, 2018.

\bibitem{He2019FurnitureFree}
Z.~He, J.~Hou, and S.~Schwertfeger, ``Furniture free mapping using 3d lidars,''
  in \emph{2019 IEEE International Conference on Robotics and Biomimetics
  (ROBIO)}, IEEE.\hskip 1em plus 0.5em minus 0.4em\relax IEEE, 2019.

\bibitem{feng2023Floorplannet}
D.~Feng, Z.~He, J.~Hou, S.~Schwertfeger, and L.~Zhang, ``Floorplannet: Learning
  topometric floorplan matching for robot localization,'' in \emph{2023
  International Conference on Robotics and Automation (ICRA)}, 2023.

\bibitem{xie2023robust}
\BIBentryALTinterwordspacing
F.~Xie and S.~Schwertfeger, ``Robust lifelong indoor lidar localization using
  the area graph,'' in \emph{arXiv}, 2023. [Online]. Available:
  \url{https://arxiv.org/abs/2308.05593}
\BIBentrySTDinterwordspacing

\bibitem{semanticslam}
E.~Michael, T.~Summers, T.~A. Wood, C.~Manzie, and I.~Shames, ``Probabilistic
  data association for semantic slam at scale,'' in \emph{2022 IEEE/RSJ
  International Conference on Intelligent Robots and Systems (IROS)}, 2022, pp.
  4359--4364.

\bibitem{QuadricSLAM}
L.~Nicholson, M.~Milford, and N.~Sünderhauf, ``Quadricslam: Dual quadrics from
  object detections as landmarks in object-oriented slam,'' \emph{IEEE Robotics
  and Automation Letters}, vol.~4, no.~1, pp. 1--8, 2019.

\bibitem{SemanticFusion}
J.~McCormac, A.~Handa, A.~Davison, and S.~Leutenegger, ``Semanticfusion: Dense
  3d semantic mapping with convolutional neural networks,'' in \emph{2017 IEEE
  International Conference on Robotics and Automation (ICRA)}, 2017, pp.
  4628--4635.

\bibitem{SemanticKITTI}
J.~Behley, M.~Garbade, A.~Milioto, J.~Quenzel, S.~Behnke, C.~Stachniss, and
  J.~Gall, ``Semantickitti: A dataset for semantic scene understanding of lidar
  sequences,'' in \emph{Proceedings of the IEEE/CVF International Conference on
  Computer Vision (ICCV)}, October 2019.

\bibitem{denseslam}
K.~Tateno, F.~Tombari, and N.~Navab, ``Real-time and scalable incremental
  segmentation on dense slam,'' in \emph{2015 IEEE/RSJ International Conference
  on Intelligent Robots and Systems (IROS)}, 2015, pp. 4465--4472.

\bibitem{Kimera}
A.~Rosinol, M.~Abate, Y.~Chang, and L.~Carlone, ``Kimera: an open-source
  library for real-time metric-semantic localization and mapping,'' in
  \emph{2020 IEEE International Conference on Robotics and Automation (ICRA)},
  2020, pp. 1689--1696.

\bibitem{KUIPERS1978129}
B.~Kuipers, ``Modeling spatial knowledge,'' \emph{Cognitive Science}, vol.~2,
  no.~2, pp. 129--153, 1978.

\bibitem{RUIZSARMIENTO2017257}
J.-R. Ruiz-Sarmiento, C.~Galindo, and J.~Gonzalez-Jimenez, ``Building
  multiversal semantic maps for mobile robot operation,'' \emph{Knowledge-Based
  Systems}, vol. 119, pp. 257--272, 2017.

\bibitem{Multi-hierarchical}
C.~Galindo, A.~Saffiotti, S.~Coradeschi, P.~Buschka, J.~Fernandez-Madrigal, and
  J.~Gonzalez, ``Multi-hierarchical semantic maps for mobile robotics,'' in
  \emph{2005 IEEE/RSJ International Conference on Intelligent Robots and
  Systems}, 2005, pp. 2278--2283.

\bibitem{3DSceneGraph}
I.~Armeni, Z.-Y. He, J.~Gwak, A.~R. Zamir, M.~Fischer, J.~Malik, and
  S.~Savarese, ``3d scene graph: A structure for unified semantics, 3d space,
  and camera,'' in \emph{Proceedings of the IEEE/CVF International Conference
  on Computer Vision (ICCV)}, October 2019.

\bibitem{rosinol20203d}
A.~Rosinol, A.~Gupta, M.~Abate, J.~Shi, and L.~Carlone, ``3d dynamic scene
  graphs: Actionable spatial perception with places, objects, and humans,''
  \emph{arXiv preprint arXiv:2002.06289}, 2020.

\bibitem{hughes2022hydra}
N.~Hughes, Y.~Chang, and L.~Carlone, ``Hydra: a real-time spatial perception
  system for 3d scene graph construction and optimization,'' 2022.

\bibitem{bormann2016room}
R.~Bormann, F.~Jordan, W.~Li, J.~Hampp, and M.~H{\"a}gele, ``Room segmentation:
  Survey, implementation, and analysis,'' in \emph{2016 IEEE international
  conference on robotics and automation (ICRA)}.\hskip 1em plus 0.5em minus
  0.4em\relax IEEE, 2016, pp. 1019--1026.

\bibitem{kleiner2017solution}
A.~Kleiner, R.~Baravalle, A.~Kolling, P.~Pilotti, and M.~Munich, ``A solution
  to room-by-room coverage for autonomous cleaning robots,'' in \emph{2017
  IEEE/RSJ International Conference on Intelligent Robots and Systems
  (IROS)}.\hskip 1em plus 0.5em minus 0.4em\relax IEEE, 2017, pp. 5346--5352.

\bibitem{liu2018floornet}
C.~Liu, J.~Wu, and Y.~Furukawa, ``Floornet: A unified framework for floorplan
  reconstruction from 3d scans,'' in \emph{Proceedings of the European
  conference on computer vision (ECCV)}, 2018, pp. 201--217.

\bibitem{wurm2008coordinated}
K.~M. Wurm, C.~Stachniss, and W.~Burgard, ``Coordinated multi-robot exploration
  using a segmentation of the environment,'' in \emph{2008 IEEE/RSJ
  International Conference on Intelligent Robots and Systems}.\hskip 1em plus
  0.5em minus 0.4em\relax IEEE, 2008, pp. 1160--1165.

\bibitem{thrun1998learning}
S.~Thrun, ``Learning metric-topological maps for indoor mobile robot
  navigation,'' \emph{Artificial Intelligence}, vol.~99, no.~1, pp. 21--71,
  1998.

\bibitem{setalaphruk2003robot}
V.~Setalaphruk, A.~Ueno, I.~Kume, Y.~Kono, and M.~Kidode, ``Robot navigation in
  corridor environments using a sketch floor map,'' in \emph{Proceedings 2003
  IEEE International Symposium on Computational Intelligence in Robotics and
  Automation. Computational Intelligence in Robotics and Automation for the New
  Millennium (Cat. No. 03EX694)}, vol.~2.\hskip 1em plus 0.5em minus
  0.4em\relax IEEE, 2003, pp. 552--557.

\bibitem{Friedman2007VoronoiRF}
S.~Friedman, H.~M. Pasula, and D.~Fox, ``Voronoi random fields: Extracting
  topological structure of indoor environments via place labeling,'' in
  \emph{International Joint Conference on Artificial Intelligence}, 2007.

\bibitem{stekovic2021montefloor}
S.~Stekovic, M.~Rad, F.~Fraundorfer, and V.~Lepetit, ``Montefloor: Extending
  mcts for reconstructing accurate large-scale floor plans,'' in
  \emph{Proceedings of the IEEE/CVF International Conference on Computer
  Vision}, 2021, pp. 16\,034--16\,043.

\bibitem{zheng2021research}
T.~Zheng, Z.~Duan, J.~Wang, G.~Lu, S.~Li, and Z.~Yu, ``Research on distance
  transform and neural network lidar information sampling classification-based
  semantic segmentation of 2d indoor room maps,'' \emph{Sensors}, vol.~21,
  no.~4, p. 1365, 2021.

\bibitem{hou2018topological}
J.~Hou, Y.~Yuan, and S.~Schwertfeger, ``Topological area graph generation and
  its application to path planning,'' \emph{arXiv preprint arXiv:1811.05113},
  2018.

\end{thebibliography}

\end{document}